\newif\iftrackchanges
\newif\iffigures
\newif\iflinenumbers

\figurestrue \trackchangesfalse \linenumbersfalse

\documentclass[10pt,letterpaper,twoside]{article}
\usepackage[top=0.85in,left=2.75in,footskip=0.75in]{geometry}

\usepackage{amsmath,amssymb,mathtools}
\usepackage[per-mode=reciprocal-positive-first]{siunitx}
\usepackage{interval, nicefrac, bm}
\usepackage[version=4]{mhchem}
\intervalconfig{soft open fences} \usepackage{mathrsfs}

\usepackage{changepage}

\usepackage[utf8]{inputenc}

\usepackage{anyfontsize}

\usepackage{textcomp,marvosym}

\usepackage{cite}

\usepackage{nameref,hyperref}
\usepackage[capitalize]{cleveref}
\crefname{figure}{Fig}{Figs}

\iffigures
\graphicspath{{./figures/}}
\else
\renewcommand\includegraphics[2][blank]{}
\makeatletter
\renewcommand*{\fps@figure}{h!}
\makeatother
\fi

\usepackage[right]{lineno}

\usepackage{microtype}
\DisableLigatures[f]{encoding = *, family = * }

\usepackage[table]{xcolor}

\iftrackchanges
\newcommand\edited[1]{\textcolor{purple}{#1}}
\else
\let\edited\relax
\fi

\usepackage{array}

\usepackage{multirow,booktabs}

\newcolumntype{+}{!{\vrule width 2pt}}

\newlength\savedwidth

\raggedright
\usepackage[aboveskip=1pt,labelfont=bf,labelsep=period,justification=raggedright,singlelinecheck=off]{caption}

\makeatletter
\renewcommand{\@biblabel}[1]{\quad#1.}
\makeatother

\usepackage{lastpage,fancyhdr,graphicx}
\usepackage{epstopdf}
\fancyheadoffset[L]{2.25in}
\fancyfootoffset[L]{2.25in}
\newcommand\term\textit
\newcommand{\celegans}{\term{C.~elegans}}

\newcommand*\dd{\mathrm{d}}
\newcommand*\diff[2][]{\frac{\dd#1}{\dd#2}}
\newcommand*\diffn[3][]{\frac{\dd^{#3}#1}{\dd{#2}^{#3}}}
\newcommand*\tdiff[2][]{\tfrac{\dd#1}{\dd#2}}
\DeclarePairedDelimiter\abs{\lvert}{\rvert}
\DeclarePairedDelimiter\norm{\lVert}{\rVert}
\makeatletter
\newcommand{\eval}{\@ifstar{\@evalar}{\@eval}}
\newcommand{\@evalar}[3]{\ensuremath{\left. #1 \right|_{#2}^{#3}}}
\newcommand{\@eval}[4][]{#2 #1|_{#3}^{#4}}
\makeatother

\begin{document}
\vspace*{0.2in}

\begin{flushleft}
{\Large
\textbf\newline{Spiking neural state machine for gait frequency entrainment
in a flexible modular robot}
}
\newline
\\
Alex Spaeth\textsuperscript{1,2,*},
Maryam Tebyani\textsuperscript{1},
David Haussler\textsuperscript{2,3},
Mircea Teodorescu\textsuperscript{1,2},
\\
\bigskip
\textbf{1} Department of Electrical and Computer Engineering,
University of California, Santa Cruz, Santa Cruz, California, United States
\\
\textbf{2} Genomics Institute,
University of California, Santa Cruz, Santa Cruz, California, United States
\\
\textbf{3} Howard Hughes Medical Institute,
University of California, Santa Cruz, Santa Cruz, California, United States
\\
\bigskip

* atspaeth@ucsc.edu

\end{flushleft}
\section*{Abstract}
We propose a modular architecture for neuromorphic closed-loop
control based on bistable relaxation oscillator modules consisting
of three spiking neurons each. Like its biological prototypes, this
basic component is robust to parameter variation but can be modulated
by external inputs. By combining these modules, we can construct a
neural state machine capable of generating the cyclic or
repetitive behaviors necessary for legged locomotion. A concrete
case study for the approach is provided by a modular robot constructed
from flexible plastic volumetric pixels, in which we produce a
forward crawling gait entrained to the natural frequency of the
robot by a minimal system of twelve neurons organized into four modules.

\iflinenumbers
\linenumbers
\fi

\section{Introduction}

\edited{
    Artificial intelligence has in recent years seen a revolution in the
    area of ``neuromorphic computation'', the use of brain-inspired
    computational units which, like biological neurons, communicate with
    each other using discrete events called ``spikes'' to perform a variety
    of tasks \cite{roy2019spikebased}. These techniques can
    have significant advantages in terms of power consumption and the
    ability to make use of the temporal structure of input
    data~\cite{abderrahmane2020design}.
}

\edited{
    It is common to derive spiking neural networks directly from deep
    learning approaches, either by modifying the learning technique to be
    applicable to a spiking neural network, or by globally converting a
    traditional neural net into an equivalent spiking neural
    network \cite{diehl2016conversion}. Both of these approaches have
    shown promise on many problems, including the generation of complex
    behaviors for simple robots \cite{milde2017obstacle, lobov2020spatial,
    bing2020indirect}. In this paper, we aim to illustrate a very
    different approach, where spiking neurons are used as elements to
    design a circuit which implements a desired behavior.
}

\edited{
    The function of interest in our application is a central pattern
    generator, or CPG, inspired by the way in which the simplest organisms
    such as nematodes use neurons to implement periodic motor patterns
    and to make simple decisions such as retreating in response to a
    toxic chemical stimulus \cite{sterling2015principles}. Pattern
    generation is a well-studied topic in computational neuroscience
    \cite{marder1996principles},
    but in robotics, central pattern generators constructed from
    spiking neurons are relatively rare. Instead, applications
    typically use the abstract concept of a CPG, but construct
    periodic motions from periodic primitives such as phase oscillators
    to produce motion commands \cite{yu2014survey}. However, an approach
    informed by neuroscience is worthwhile because robotic
    experiments in this vein can generate interesting insights about the
    design tradeoffs made by real neural systems \cite{webb2020robots}.
}

Applications of this principle to robotics include a variety of open-loop
spiking CPGs generating position commands for robots ranging from
hexapods \cite{strohmer2020flexible} to fish
\cite{korkmaz2020locomotion}. Also, certain researchers have decided
to tackle the problem of proprioceptive feedback in quadrupeds
\cite{maruyama2015hardwired, hunt2015biologically, habu2018simple}.
In this paper, which extends work previously presented at the 2020
IEEE International Conference on Soft
Robotics~\cite{spaeth2020neuromorphic} using a prototype of the same
robot, we will demonstrate the design and implementation of a spiking
central pattern generator entrained by proprioceptive feedback to
control the locomotion of a modular robot. This system serves as a
case study for a simple conceptual framework which allows us to
implement closed-loop robotic control through a minimal spiking neural
network without recourse to black-box optimization.

The organization of this paper is as follows. \Cref{s:blocks}
describes the basic module making up the controller as well as
how such modules can be used to implement arbitrary state
machines. \Cref{s:robustness} demonstrates that the computational
properties of this module are insensitive to parameter variation yet
relatively easy to modulate with the appropriate inputs. \Cref{s:casestudy}
applies these principles to generate a specific desired state machine for
the control of a flexible robot. Finally, \cref{s:physical} describes the
experiments that were carried out on this platform.

\section{Building Blocks for Universal Computation}
\label{s:blocks}

\edited{
    From a design perspective, it can be helpful to treat neurons as
    asynchronous digital logic elements which represent an \textsc{on}
    state by firing a train of action potentials at a given rate, and
    \textsc{off} by remaining quiescent. This simplified model necessarily
    ignores many subtleties of the underlying analog neural system, and
    it is not Turing complete \cite{sima2020analog}, but it can be used
    to implement neural state machines for robotic control \cite{liang2019neural}.
    }

\begin{figure}
    \centering
    \includegraphics{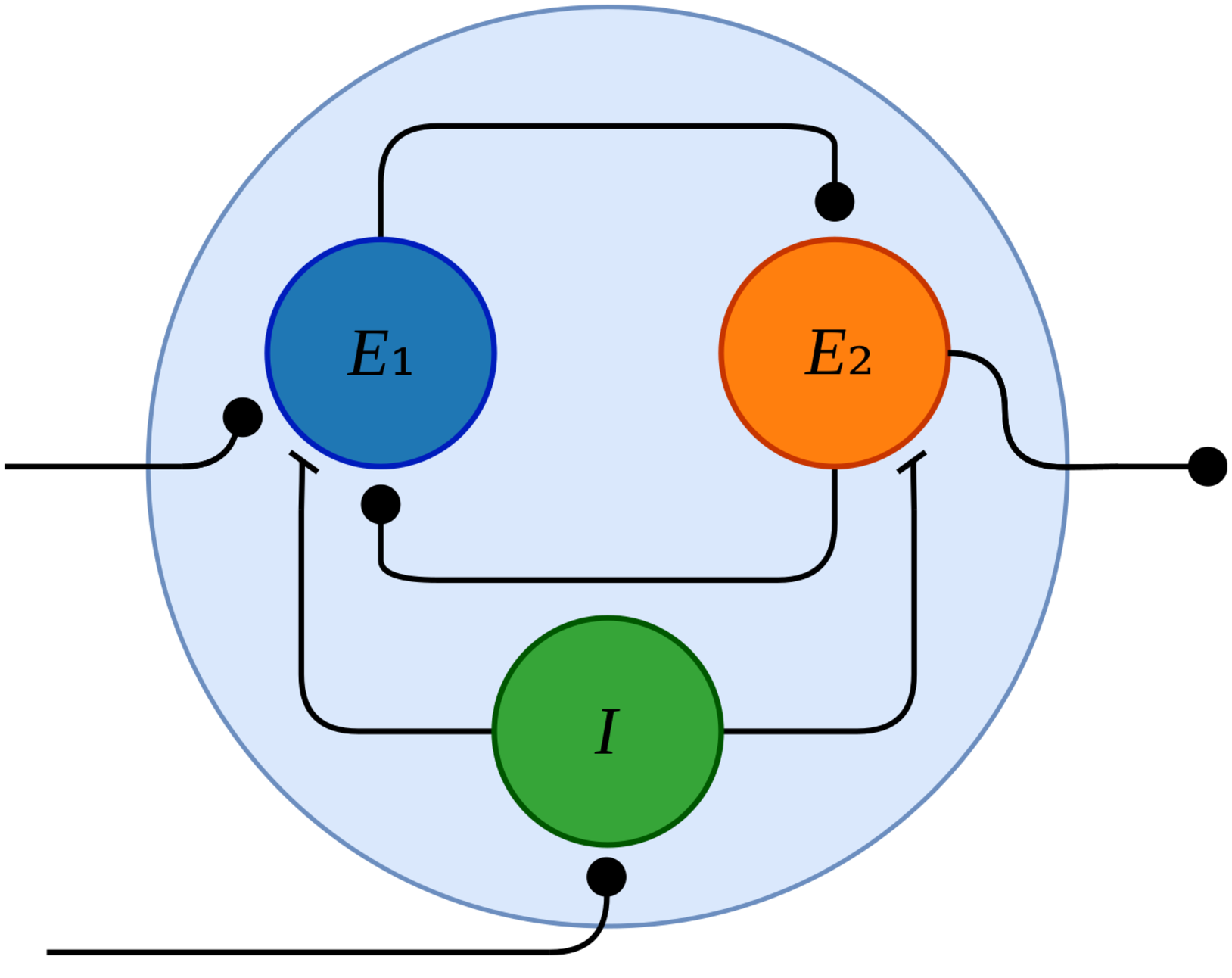}
    \caption{\textbf{The neural latch module.} Arrowheads are circular
        (flat) to represent excitatory (resp.\ inhibitory)
        connections. The symmetrically connected pair of excitatory
        neurons $E_1$ and $E_2$ are connected to produce positive
        feedback. An excitatory input to $E_1$ on the left activates
        the module, and the output is taken from $E_2$ on the right.
        The inhibitory interneuron $I$ can deactivate the module in
        response to an external signal which enters on the bottom
        left.
    } \label{f:oscillator}
\end{figure}

\edited{In this section, we propose a basic module which can be used to
construct such a state machine in the setting of neuromorphic computation:
a three-neuron circuit which can hold state, depicted in
\cref{f:oscillator}. This module is a  neuromorphic implementation of an
electronic set-reset (SR) latch, a component which holds one bit of state
controlled by its two inputs, ``set'' (S) and ``reset'' (R). Like its
electronic prototype, this circuit} can switch between two
states depending on its input: an inactive resting state representing
a saved logical value of 0, and an active spiking limit cycle
representing a logical 1. We can construct such a circuit by
connecting two neurons, \edited{labeled $E_1$ and $E_2$ in the figure},
in strong mutual excitation; in this
configuration, whenever either neuron fires, the other will fire in
response, but the system can remain in a quiescent state for any
duration if neither neuron is externally induced to fire.

\edited{
    Technically the described behavior is not tonic firing, although
    the time series of spikes appears that way. Tonic firing occurs
    when a neuron fires repeatedly in response to a constant current
    or due to its own intrinsic dynamics, whereas the neurons in this
    module only produce single action potentials in response to
    presynaptic firings. In that sense, this module is an oscillator
    constructed from two neurons each individually incapable of
    oscillation. Similarly, when the module state is toggled by outside
    stimuli, its behavior may appear superficially similar to bursting,
    but the underlying dynamics are fundamentally  different
    \cite{izhikevich2006dynamical}. This mechanism of rhythmogenesis
    is certainly not typical for biological central pattern generation
    \cite{strohmer2020flexible}, but it is effective from an engineering
    perspective.
}

Biological neurons are classified broadly as either excitatory or inhibitory,
and inhibitory interactions are typically short range, so to complete the
circuit we introduce a local inhibitory interneuron $I$ within the module,
which we call the ``reset interneuron''. This neuron, which provides the
only inhibitory inputs to $E_1$ and $E_2$, can receive a long-range excitatory
connection to function as the ``reset'' input in the SR latch metaphor.
Weak excitatory synapses from $E_1$ and $E_2$ to $I$
allow the latch to modulate its own reset: an external input just
barely strong enough to trigger a reset while the latch is active will
not trigger a reset while the latch is inactive, conserving the energy
of the interneuron.

\begin{figure}
    \centering
    \includegraphics{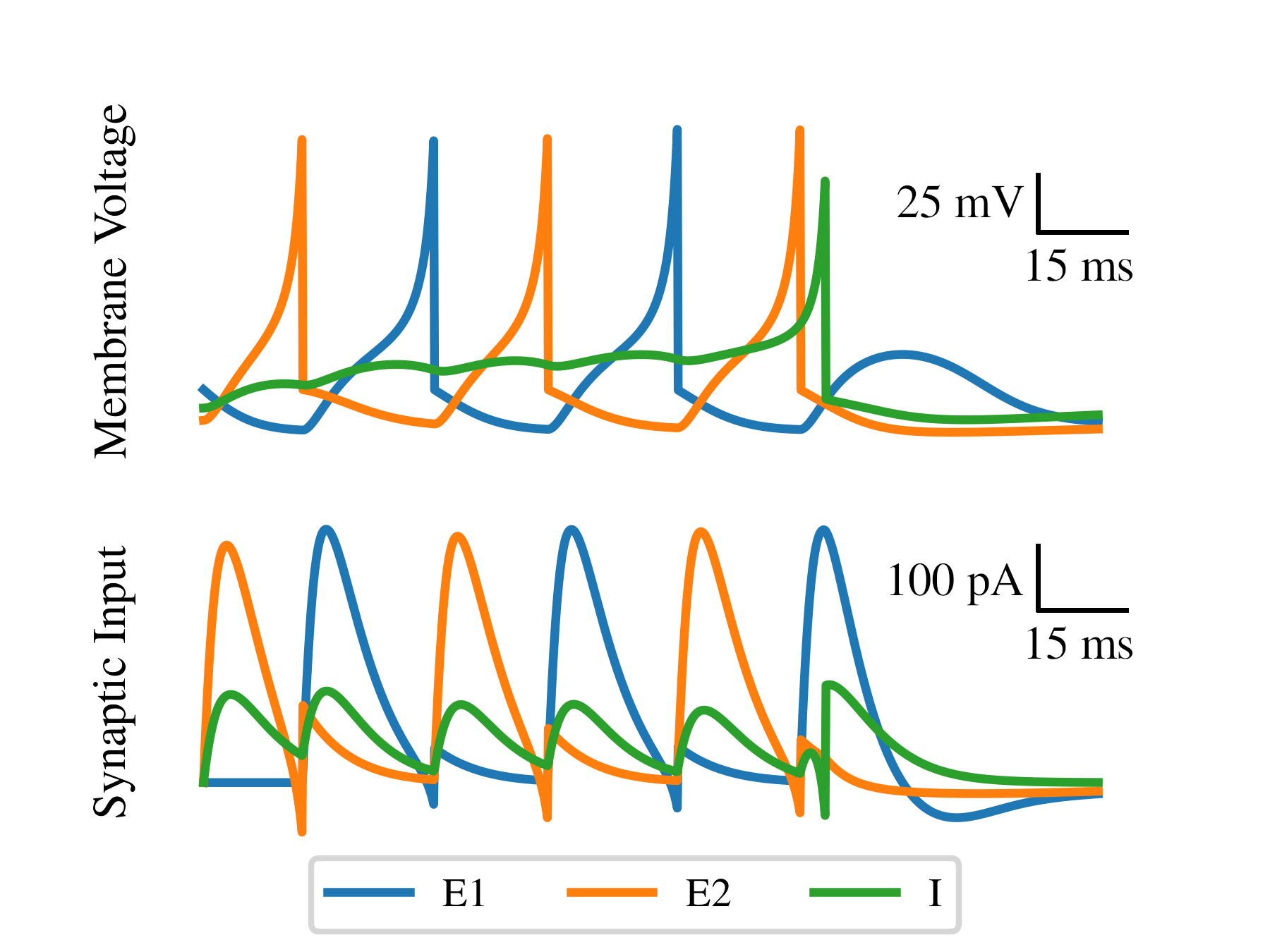}
    \caption{\textbf{Activity in the spiking neural module.}
        \edited{Membrane voltage and input synaptic current for each
        neuron are displayed as the module} is deactivated by some
        external stimulus. The three neurons are represented by their
        membrane voltages, with three traces corresponding to the
        color code established in \cref{f:oscillator}: the two
        excitatory neurons are shown in blue and orange, while the
        inhibitory neuron is in green.
    } \label{f:openloop}
\end{figure}

\edited{
The activity of the three neurons in the neural module is depicted in
\cref{f:openloop}. The excitatory neurons $E_1$ and $E_2$ fire in
alternation until an externally-induced firing of the reset
interneuron $I$ silences them.
    Note that each excitatory neuron fires only when the other
    provides enough synaptic current to activate it. Activity stops
    because the final pulse of synaptic input from $E_2$ to $E_1$ is
    curtailed by the single firing of $I$. The effect of this
    inhibition is not obvious in the plot of synaptic currents due to
    the slow time constant of inhibition in this model; it is visible
in the shape of the tail more so than in the height of the peak.
However, even this small change is sufficient to stop $E_1$ from firing,
deactivating the module.}

\subsection{Breaking the Digital Mold}

Although it can help in the conceptual design phase to view neural
systems through a digital lens, such models can be overly limiting in
terms of efficiency and expressiveness. Many functions which would
require a very large number of logic gates to implement using binary
neurons can be realized in a relatively straightforward way using the
analog properties of a much smaller number of spiking neurons.

From an electrophysiological perspective, a neuron is an electrically
active cell with a variety of ionic channels which maintain the
membrane voltage near a resting value but can generate large
spike-shaped excursions called ``action potentials'' in response to
currents induced by input neurotransmitters. Spiking neural models are
those which reproduce this excitability in the form of an excitable
dynamical system.

A good intuitive model of this behavior is given by the leaky
integrate-and-fire (LIF) neuron, in which the membrane voltage $v$ has
linear dynamics tending towards a fixed resting potential $V_r$, but
if the collective effect of the inputs drives the membrane voltage
above the peak value $V_p$, a spiking event is said to have occurred,
and the membrane voltage is reset to a fixed value $c$.

As an example of how these dynamics may be useful, the simplest
realization of a delay in a digital circuit is a series of many
identical short delay elements such as inverters, whereas an analog
system can implement a delay through parameter tuning without
introducing new circuit elements. A small excitatory input can
gradually increase the membrane voltage of a neuron until it fires,
\edited{so that a sequence of many presynaptic firings at sufficiently
high frequency is required} before any signal is propagated to the output.
In fact, it is exactly this method which we will use in
\cref{s:parameters} to ensure that the activation of each neural
module propagates only gradually to other modules to which it is
connected.

\edited{Instead of viewing the neural module within the digital logic
metaphor, we can take the dynamical perspective and treat it as a
bistable relaxation oscillator. If a neuron does
not fire, it does not contribute to the positive feedback loop
responsible for the oscillatory state, so} the state of the module as
a whole remains unchanged for subthreshold inputs. On the other hand,
for inputs large enough to produce a spike, there is positive
closed-loop gain, producing oscillatory behavior.

It is worth noting that in the context of continuous dynamics, the
analogy to a state machine is made by a so-called ``heteroclinic
network'', where the active limit cycle of a module represents an
attractor corresponding to a state \cite{egbert2020where}. Inputs,
disturbances, or weak coupling between modules can move the system
state from one attractor to another, producing a state transition
along a heteroclinic trajectory.

Other dynamical properties of the neuron may also be used for
computation. The morphological diversity of the real neuron seems to
be used to great advantage in this context: dendritic computation and
the availability of a wide array of different neurotransmitters and
classes of inhibition can allow the implementation of logic gates at a
finer granularity than the individual cell
\cite{sterling2015principles}. \edited{Likewise, it is possible for
the dynamics of individual neurons to produce bistability between
resting and tonic firing or between tonic firing and bursting
\cite{shilnikov2005mechanism}.} Additionally, besides the
``integrator-type'' neurons which we have described conceptually,
neurons also exist which are electrically resonant, responding
preferentially to inputs which come at a certain frequency
\cite{izhikevich2006dynamical}. However, we will not be exploring
these phenomena in this paper.

\subsection{Modeling Spiking Neurons}

In order to take advantage of the analog computational properties of
spiking neurons, we apply the Izhikevich model together with
conductance-based synapses to describe the behavior of our simulated
neurons. We have chosen to adapt Izhikevich's spiking neural model
because it trades off phenomenological accuracy with efficient
computation in a reasonable way \cite{izhikevich2004which}.
This model is essentially a
LIF neuron augmented with two main features: a quadratic nonlinearity
to reproduce the dynamics of the biological threshold behavior, and
the recovery variable $u$, which abstracts various slow ionic currents
to provide persistent state across spikes.  The model parameters
\edited{are physically interpretable,} and the model dynamics can be
fit to observed electrophysiological data.

Although the Izhikevich model is not the simplest neural model capable
of producing a bistable oscillator network \cite{ermentrout1996type},
its accurate recapitulation of electrophysiological activity is
important from a design perspective. Qualitatively similar behavior
can be generated by a variety of different systems, but greater
fidelity to biological prototypes helps a system designer consider
constraints comparable to those faced in the evolution
of biological organisms.

Once individual neurons have been modeled, the next consideration is
how to connect them in networks. Large neural networks commonly employ
a Dirac delta synapse, where each firing of a presynaptic neuron
instantaneously increases the membrane voltage of the postsynaptic
neuron. However, phase synchronization of integrator-type neurons
requires synaptic activity with finite nonzero duration
\cite{izhikevich1999class}. For this reason, we implement
conductance-based synapses, where each synaptic connection is
represented as a conducting channel gated by a presynaptic activation
$x$, which increases after a firing event and gradually decays back to
zero. When activated, the channel admits a current proportional to the
presynaptic activation times the difference between the postsynaptic
membrane potential and the synaptic reversal potential
\cite{hille1992ion}.

The time course of the presynaptic activation follows the classic
``alpha'' function $x(t) = \frac{t}{\tau} e^{-t/\tau}$ observed for
synaptic conductances \term{in vivo} \cite{rall1967distinguishing}.
This function is the solution to the second-order linear differential
equation $\tau^2\diffn[x]t2 + 2\tau\diff[x]t + x = 0$, with the
initial conditions $x(0) = 0$ and $\tau\eval{\diff[x]t}{t=0}{} = 1$,
so separating into two first-order differential equations by
introducing a second variable $y = \tau\diff[x]{t}$, we obtain $x$ and
$y$ dynamics with which the Izhikevich model can be augmented to
produce the following dynamical model of a single neuron subject to an
input current $I(t)$:

\begin{equation}\label{eq:izhikevich}
    \begin{aligned}
        \text{When $v < v_p$}
        && C\tdiff{t}v &= k(v - v_r)(v - v_t)
        - u + I(t) \\
        && \tdiff{t}u &= a(b(v - v_r) - u)\\
        && \tdiff{t}x &= j / \tau\\
        && \tdiff{t}y &= -(2y + x) / \tau\\
        \text{When $v = v_p$}
        &&  v &\gets c \\
        &&  u &\gets u + d\\
        &&  y &\gets y + 1
    \end{aligned}
\end{equation}

After equipping these neurons with the parameters described in
\cref{t:parameters}, we can combine them into neural networks
consisting of $N$
neurons coupled to each other only through the input current $I(t)$.
The magnitude of this current is parameterized by a matrix $G$ whose
entries $g_i^j$ specify the peak input synaptic conductance
triggered in the $i$th neuron by a firing in the $j$th neuron. The
synaptic reversal potential $V_n$ is also specified per presynaptic
neuron. The resulting input $I_i(t)$ to the $i$th neuron is therefore
given by:

\begin{equation}\label{eq:input-current}
    I_i(t) = \sum_{j=1}^N g_i^j x_j (V_{n,j} - v_i)
\end{equation}

\begin{table}
  \centering
  \caption{\textbf{Parameters for the Izhikevich neuron.}}
  \begin{tabular}{clccc}\toprule
    \multicolumn{2}{c}{\multirow{2}{*}{Parameter}}
    & \multicolumn{2}{c}{Cell Type} & \multirow{2}{*}{Units} \\
    \cmidrule{3-4}
    & & RS & LTS & \\
    \midrule
    $a$ & characteristic rate of $u$ & 0.03 & 0.03 & \si{\per\ms}\\
    $b$ & leakage conductance & -2 & 8 & \si{\nano\siemens} \\
    $c$ & downstroke return voltage & -50 & -53 & \si{\mV} \\
    $d$ & downstroke inrush current & 100 & 20 & \si{\pA} \\
    $C$ & membrane capacitance & 100 & 100 & \si{\pF} \\
    % Put the unit in a zero-width box to suppress overfull hbox.
    $k$ & sodium channel gain & 0.7 & 1.0 &
    \makebox[0em][c]{\si{\nano\siemens\per\mV}} \\
    $V_r$ & resting potential & -60 & -56 & \si{\mV} \\
    $V_t$ & threshold voltage & -40 & -42& \si{\mV} \\
    $V_n$ & synaptic reversal potential & 0 & -70 & \si{\mV} \\
    $V_p$ & action potential peak & 35 & 20 & \si{\mV}  \\
    $\tau$ & synaptic time constant & 5 & 20 & \si{\ms} \\\bottomrule
  \end{tabular}
  \label{t:parameters}
\end{table}

Finally, we generate an actuation effort command for the physical
actuators using simple simulated muscle cells. Because our system is
designed with invertebrate control schemes in mind, it is reasonable
to use a muscle model where, unlike the electrically active vertebrate
muscle cell, muscle cells do not fire action potentials. Instead, they
simply provide contractile force in proportion to their depolarization
\cite{naris2020neuromechanical}. The result is similar to applying an
exponential low-pass filter to the spiking waveform
\cite{strohmer2020flexible}, smoothing the
output of the system so that changes occur only on the timescale of a
few milliseconds, determined by the synaptic and membrane time
constants.

\subsection{Choosing Model Parameters} \label{s:parameters}

We have designed the neural latch of \cref{f:oscillator} according
to a simplified binary neural model, whereas the spiking dynamical
neural model with which we are simulating our circuit requires all of
the various parameters summarized in \cref{t:parameters}. The
question then arises of how one should choose the values of these
parameters in order to obtain some desired behavior.

We use the parameter values shown in \cref{t:parameters},
corresponding to well-characterized human cell types from the
literature \cite{izhikevich2006dynamical}. Excitatory cells are
modeled as regular spiking (RS) pyramidal cells, and inhibitory cells
take parameters corresponding to low-threshold spiking (LTS)
inhibitory interneurons. \edited{These parameters have been chosen
mainly based on their ready availability and the fact that they do not
have intrinsic bursting dynamics or bistability, allowing us to
implement the essential features of the module in circuitry rather
than depending on exotic dynamics. However, it is fully possible to
use parameters corresponding to other cell types in order to implement
the same behaviors. For example, if we had used a set of parameters
which produces bursting dynamics, our module might have implemented a
limit cycle where the two neurons alternate bursts rather than spikes,
but the same logical operations and constructions would have been
available to us. As we will see later, this structural style of design
leads to robust behavior in the module, in contrast with the precise
tuning that would typically be necessary if all of our logical
operations were implemented using such analog properties.}

\edited{With these values given, only the following synaptic
conductances must be tuned manually in order to implement any desired
logical structure}: the strong connection $G_\text{exc}$ between the
two excitatory neurons within the module, another relatively strong
synapse $G_\text{inh}$ allowing the reset interneuron $I$ to deactivate
the module, \edited{a weak synapse $G_\text{rst}$ priming the reset to
fire while the module is active}, a relatively weak feedforward input
$G_\text{ffw}$ by which one module might gradually activate another,
and a moderately strong connection $G_\text{fb}$ by which one module
can quickly deactivate another. We found empirically that these
parameters can be set with only a small amount of trial-and-error fine
tuning due to their wide tolerances, \edited{as we will see in
\cref{s:poincare}. Furthermore, three of them are internal to the
module, and so robustness in these parameters is inherited by larger
networks constructed from the same building block. The values we
arrived at are given in \cref{t:synaptic-parameters}.}

\begin{table}
    \centering
    \edited{
    \caption{\edited{\textbf{Synaptic parameters for the neural latch module.}}}
    % S[table-format=3.2] is from siunitx and allows aligning numbers
    % for easy comparison by making the digits correspond
    \begin{tabular}{clS[table-format=3.2]}\toprule
        \multicolumn{2}{c}{Parameter and Use} &
        % Braces required to avoid treating as a number.
        {Value (\si{\nano\siemens})}\\
        \midrule
        $G_\text{exc}$ & connection within the module & 20 \\
        $G_\text{inh}$ & strength of the ``reset'' interneuron & 10 \\
        $G_\text{rst}$ & priming the reset during module activity & 5 \\
        \midrule
        $G_\text{ffw}$ & gradual activation of other modules & 10 \\
        $G_\text{fb}$ & deactivation of other modules & 10 \\
        \bottomrule
    \end{tabular}
    \label{t:synaptic-parameters}
}
\end{table}

The neuron parameters affect the behavior of the module, but to varying
degree. For instance, the action potential peak $V_p$ can be set almost
completely arbitrarily because by the peak of an action potential, $v$
is on a trajectory which would reach infinity in only a few milliseconds
if not for the spike reset. As a result, the choice of precisely where to
clip the spike does not substantially alter the trajectory
\cite{izhikevich2006dynamical}.

Other parameters have a non-negligible effect on the
period of the oscillation within the module, but for many of these,
since their main effect is essentially shifting the timescale on which
the neuronal dynamics play out, they have much less effect on the
quantities which we want to control, such as the activation thresholds
and necessary synaptic conductances. For example, increasing the
after-spike reset voltage $c$ essentially gives the dynamics of the
neuron a head start along the
trajectory that must be traversed between each firing, meaning
that the period decreases, but because it does not affect the
continuous dynamics, this has no effect at all on the process of
initiating the active state.

More pertinent to activation behavior, the strength of the excitation
from one neuron to another is controlled by several synaptic parameters:
the synaptic time constant $\tau$, the peak synaptic conductance
$G_\text{exc}$, and the synaptic reversal potential $V_n$ all control
the shape and scale of the postsynaptic potential initiated by a
firing event. A synaptic activation with a longer duration, higher
peak conductance, or higher synaptic reversal potential will produce
a larger or longer postsynaptic potential, and therefore cause a
postsynaptic firing either more quickly or more surely.

One may imagine the neuron as a simple circuit with a
single capacitance representing the cell membrane, and
\edited{transmembrane conductances connecting this capacitor to
voltage sources corresponding to ionic reversal potentials. In
particular, the synaptic conductance connects to the reversal
potential $V_n$. This view is the basis of biophysical models of
neuronal dynamics (see, for example, ref.~\cite{hille1992ion}), but as
a mental image it can potentially be misleading because it encourages
visualizing the steady state of an RC circuit, where the actual value
of the conductance is unimportant}; in fact, the synaptic conductance
is only nonzero for a short time relative to the RC time constant,
generating a small deviation in the voltage $v$ across the capacitor.
As a result, the value of the conductance actually has a substantial
impact on the amplitude of the postsynaptic potential, directly
affecting whether or not the membrane voltage surpasses the threshold.
Beyond this minimum, there is a wide range of acceptable values for
this conductance.

\section{Robustness of the Neural Module} \label{s:robustness}

Biological central pattern generation is carried out by small networks
of neurons which must be robust to certain types of parameter
variation in order for their behavior to withstand changes in the
environment, but at the same time must be sensitive to other changes
so that their behavior can be modulated by other components of the
nervous system \cite{marder2014neuromodulation}. Although certain
organisms may utilize the redundancy of large cell populations in
neural circuits to mitigate unavoidable variations in neuronal
parameters \cite{donati2014spiking}, our constructed neural system is
not redundant and so must be intrinsically robust.

In this section, we quantify the behavior of the neural latch module
of \cref{f:oscillator} with respect to variations in the parameters of
the two excitatory neurons. First, dynamical analysis is applied to
measure the maximum deviation in each parameter for which the
existence of the spiking limit cycle is preserved. Next, we use a
Monte Carlo approach to quantify the sensitivity of the module to
simultaneous variation in all twelve parameters of both neurons.

\subsection{Preservation of the Spiking Limit Cycle} \par

\label{s:poincare}

\edited{
    The essential function of the module is to switch between resting
    and spiking states in response to external input. In the absence
    of bounds on the strength of the input, as long as both states
    exist and continue to be attractive, it is always possible to
    switch between them. Therefore, we operationalize the question of
    whether a given parameter set produces a viable module by checking
    for the existence and stability of both attractors.
}

The existence of the resting state can be checked by finding a fixed
point of the dynamics, where all nullclines in the system intersect;
since three of the four phase variables of each neuron obey linear
dynamics, a fixed point can exist only when both neurons have $(u, x, y)
= (b(v-V_r), 0, 0)$. Under this condition, since the synaptic
activation variable $x$ is zero, the neurons are decoupled, and the
quadratic dynamics of the single remaining phase variable $v$ have
roots at $v = V_r$ and $v = V_t + \nicefrac{b}{k}$. Varying the
parameters $b$, $k$, $V_r$, and $V_t$ produces a transcritical
bifurcation when $\nicefrac b k = -(V_t - V_r)$. For these parameter
values, the two fixed points coincide, resulting in a single
nonhyperbolic half-stable fixed point. However, for all other values
of the parameters, the continuous dynamics exhibit exactly one stable
node separated from the firing threshold by the unstable manifold of a
saddle point, meaning that \edited{an attractive resting state always
    exists, although it is only marginally stable at a single
    pathological point in the parameter space.}

\edited{Since the resting state is always attractive, only the
existence of the attractive limit cycle must be verified for different
parameter sets.} However, it is nontrivial to prove the existence of a
limit cycle, especially for a discontinuous dynamical system such as
this one. Instead, we determine acceptable parameter values by
numerically approximating a discrete dynamical system called the
Poincaré first return map. This map is defined from an underlying
continuous dynamical system by choosing a manifold $\mathcal M$ in
phase space known to be transversal to flow. The Poincaré map $P
\colon \mathcal{M} \to \mathcal{M}$ is then defined by following the
dynamics from each point on $\mathcal{M}$ to the next point where the
flow intersects $\mathcal{M}$. Any fixed point of $P$ lies on a periodic
orbit in the original system, which is an attracting limit cycle if
the fixed point is attractive. This is explained in more detail in
any standard text on dynamical systems theory (e.g. chapter 10 of
ref.~\cite{wiggins2003introduction}).

Our Poincaré section $\mathcal{M}$ is the hyperplane $v_1 = v_p$. As a
result, the Poincaré map $P$ takes the state vector at each firing of
the first neuron to its value the next time the first neuron fires. We
compute this map numerically \edited{in the attached Julia simulation
code using the ODE solver library DifferentialEquations.jl
\cite{rackauckas2017differentialequations}}, then find its fixed point
by Picard iteration, the process of iteratively applying the
map until the results converge, starting at an arbitrary point on the
Poincaré section. Fixed points found in this manner are attractive,
meaning that they correspond to stable limit cycles. \edited{We also
checked that} if the limit cycle exists, its period is more than
\SI{5}{\ms} so that the dynamics will remain numerically stable when
computed on the robot, where we must use fixed-timestep integration
with $\Delta t$ in the hundreds of microseconds. \edited{However, this
requirement did not affect the range of acceptable parameters.}

\edited{We first check for the persistence of the spiking limit cycle
when the module is subject to variations in individual parameters, with
both excitatory neurons modulated identically. The computed acceptable
parameter ranges} are given in \cref{t:constraints}.
Parameter differences between the two neurons typically lead to
asymmetries in the spiking pattern, where the two neurons take different
amounts of time to activate, \edited{although this effect does not
interfere with the functionality of the module. Other undesirable
effects include overexcitation causing desynchronization between the
two neurons, leading to a chaotic spiking attractor rather than a
limit cycle. In principle the module could still perform its function
under these conditions, but it is so different from our intended
behavior that we consider it defective. $G_\text{exc}$, $\tau$, and
$V_n$ have their own interesting failure mode. As the firing rate
increases, for a given value of $G_\text{rst}$, the neuron can trigger
its own reset, breaking up the limit cycle. When $G_\text{rst}$ is set
to zero, these parameters can be set as large as numerical stability
will allow.}

\edited{
    Our prior statement that it is always possible to switch between
    the two attractive states if they both exist is only true if
    outside input to the module is unconstrained. However, inhibition
    to a module should come only from its own reset interneuron. The
    result is one additional constraint: the conductance
    $G_\text{inh}$ must be strong enough to move the module from the
    spiking limit cycle to the resting state. It is simple to verify
    this using the Poincaré map techniques already developed.
    Specifically, rather than ensuring that under the given parameter
    values a spiking initial condition eventually converges to the
    limit cycle, we ensure that if $G_\text{rst}$ is increased enough
    that the module triggers its own reset, the resulting synaptic
    activity is sufficient to break the limit cycle.
}

\edited{
    When this is the case, we already know that any outside input
    sufficiently strong to cause either of the neurons $E_1$ or $E_2$ to
    fire will bring the module to its spiking limit cycle.
    Furthermore, we have established that an outside input
    sufficiently strong to trigger a firing of the reset interneuron
    will bring the module back to its resting state. This process
    therefore establishes both the bistability of the system and the
    feasibility of switching between the two attractors in response to
    external input which is constrained to be excitatory.
}

\begin{table}
    \centering
    \caption{\textbf{Parameter ranges for the neural latch.}}
    \begin{tabular}{ccccc}\toprule
        Param. & Min. & Nom. & Max. & Units\\ \midrule
        $a$ & $0.017$ & $0.030$ & $\infty$ & \si{\per\ms} \\
        $b$ & $-6.08$ & $-2.0$ & $0$ & \si{\nano\siemens} \\
        $c$ & $-\infty$ & $-50.0$ & $-40.3$ & \si{\mV} \\
        $d$ & $40.4$ & $100$ & $169$ & \si{\pA} \\
        $C$ & $68.3$  & $100$ & $159$ & \si{\pF} \\
        $k$ & $0$ & $0.70$ & $1.41$
            & \si{\nano\siemens\per\mV}\hspace{-1em} \\
        $V_r$ & $-66.3$ & $-60.0$ & $-29.2$ & \si{\mV} \\
        $V_t$ & $-47.0$ & $-40.0$ & $-36.2$ & \si{\mV} \\
        $V_p$ & $-32.7$ & $35.0$ & $\infty$ & \si{\mV}  \\
        $V_n$ & $-9.7$ & $0.0$ & $9.7$ & \si{\mV} \\
        $\tau$ & $3.77$ & $5.0$ & $7.41$ & \si{\ms} \\
        \midrule
        $G_\text{exc}$ & $16.1$ & $20.0$ & $31.6$ & \si{\nano\siemens} \\
        $G_\text{rst}$ & $0.0$ & $5.0$ & $7.1$ & \si{\nano\siemens} \\
        $G_\text{inh}$ & $4.4$ & $10.0$ & $\infty$ & \si{\nano\siemens} \\
        \bottomrule
    \end{tabular}
    \label{t:constraints}
\end{table}

\subsection{Monte Carlo Analysis} \label{s:montecarlo}

The method of the previous section identifies the maximum
independently acceptable deviation in each individual parameter value,
but concurrent deviations may interact in nontrivial ways. For
example, a lower resting voltage $V_r$ or a higher threshold $V_t$
makes the resting state more attractive, but this can be compensated
by increasing the synaptic reversal potential $V_n$.

In order to generalize to simultaneous variations in all neuron
parameters, we use a Monte Carlo approach to study fractional parameter
variation. We introduce a fractional variation totaling 10\%
by selecting points $\xi$ from a 11-dimensional standard Gaussian
distribution and rescaling to set $\norm{\xi} = 1$. We then multiply
each of the parameters by 10\% of the base value of the corresponding
parameter, and add this value to the initial parameter vector. This
method selects a random parameter variation totaling exactly 10\%, but
because $\xi$ has been sampled uniformly from the surface of the unit
sphere, its components are not independent: a large deviation in one
direction typically corresponds to smaller deviations in other
parameters.

\begin{figure}
    \centering
    \includegraphics{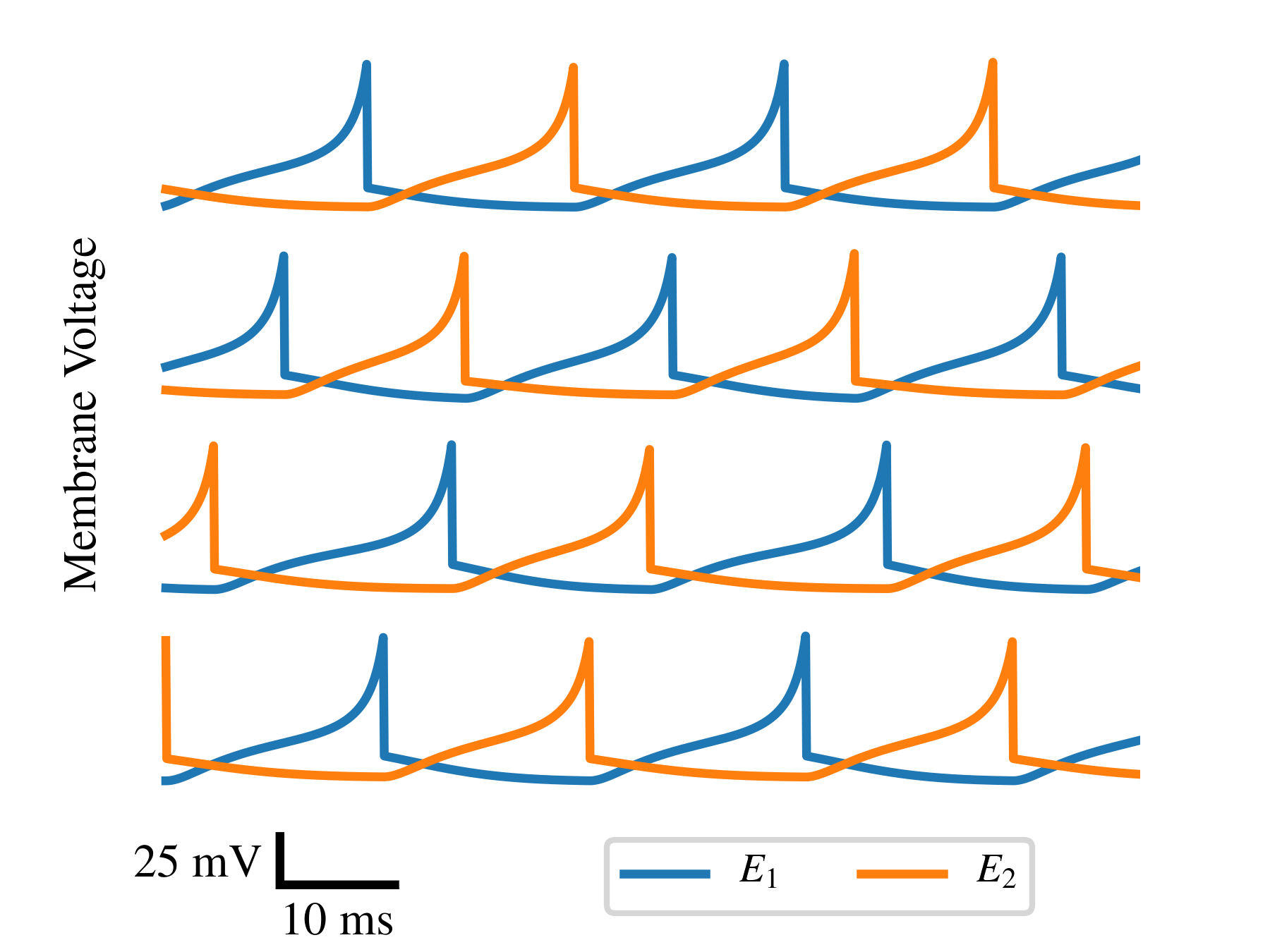}
    \caption{\textbf{Effect of random parameter variation.}
        The module was simulated four times with \edited{randomized}
        parameter values. The four plots represent the membrane
        voltage of the two excitatory neurons $E_1$ and $E_2$ in these
        four independent experiments. The top trace corresponds to the
        nominal parameter values; the lower three are the result of
        random modulation of all parameters of both neurons. The traces
        are qualitatively identical, and differ only in details of
        spike timing.
    } \label{f:parameter-variation}
\end{figure}

\edited{For illustrative purposes,} a few examples of the typical
effect of this type of parameter variation on the behavior of the
module are demonstrated in \cref{f:parameter-variation}. The membrane
voltage of the two excitatory neurons during the spiking limit cycle
of the unmodified module is compared to three different
\edited{modules, each with all parameters of both neurons $E_1$ and
$E_2$ randomly adjusted by 10\%.} In each of these cases, although the
relative timing of spiking events varies substantially, the main
qualitative features are identical: the two excitatory neurons
alternate firing in a consistent pattern.

\edited{Although \cref{f:parameter-variation} demonstrates three cases
where random parameter variation did not affect the qualitative
behavior of the module, this is not always the case. Sometimes, the
variant module is not viable in that its dynamics converge globally to
the resting state. Whether a module functions correctly or not depends
on the parameter change which occurs.} We quantified these effects
with a Monte Carlo approach, where one million random variant
\edited{modules} were tested to approximate the probability with which
an introduced 10\% parameter variation causes the module to no longer
function. This test was run in three separate variants: modifying only
one of the two neurons led to failure in \SI{1.3}{\percent} of cases,
whereas applying the same modification to both neurons led to failure
in \SI{2.8}{\percent} of cases, and modifying both neurons
independently led to failure in \SI{5.4}{\percent} of cases.

Variations in different parameters have different importance.
Many of the parameters have little effect on the observable dynamics
of the neurons when only small deviations are considered, while other
parameters may be close to a threshold beyond which the module cannot
function. To study this effect, we consider a point cloud of the
twelve-dimensional relative deviation variables $\xi$. \edited{These
are unitless random variables, generated as described above, which
specify the relative deviation in each parameter. The points in this
cloud are} labeled according to whether the module continued to
function in the sense of \cref{s:poincare} when subjected to that
deviation. \edited{Since this point cloud is symmetric, dimensionality
reductions which revolve around spatial clustering, such as principal
component analysis (PCA), are not applicable. Instead, we use our
labels for linear discriminant analysis (LDA), a simple machine
learning method based on the optimization problem of finding the
hyperplane which best separates two label classes. In this case, we
are not trying to classify our points, but we can use the hyperplane
for visualization.} The normal vector of this hyperplane is a basis
vector in the higher-dimensional space such that the projection of the
point cloud onto this axis is maximally informative.

\begin{figure}
    \centering
    \includegraphics{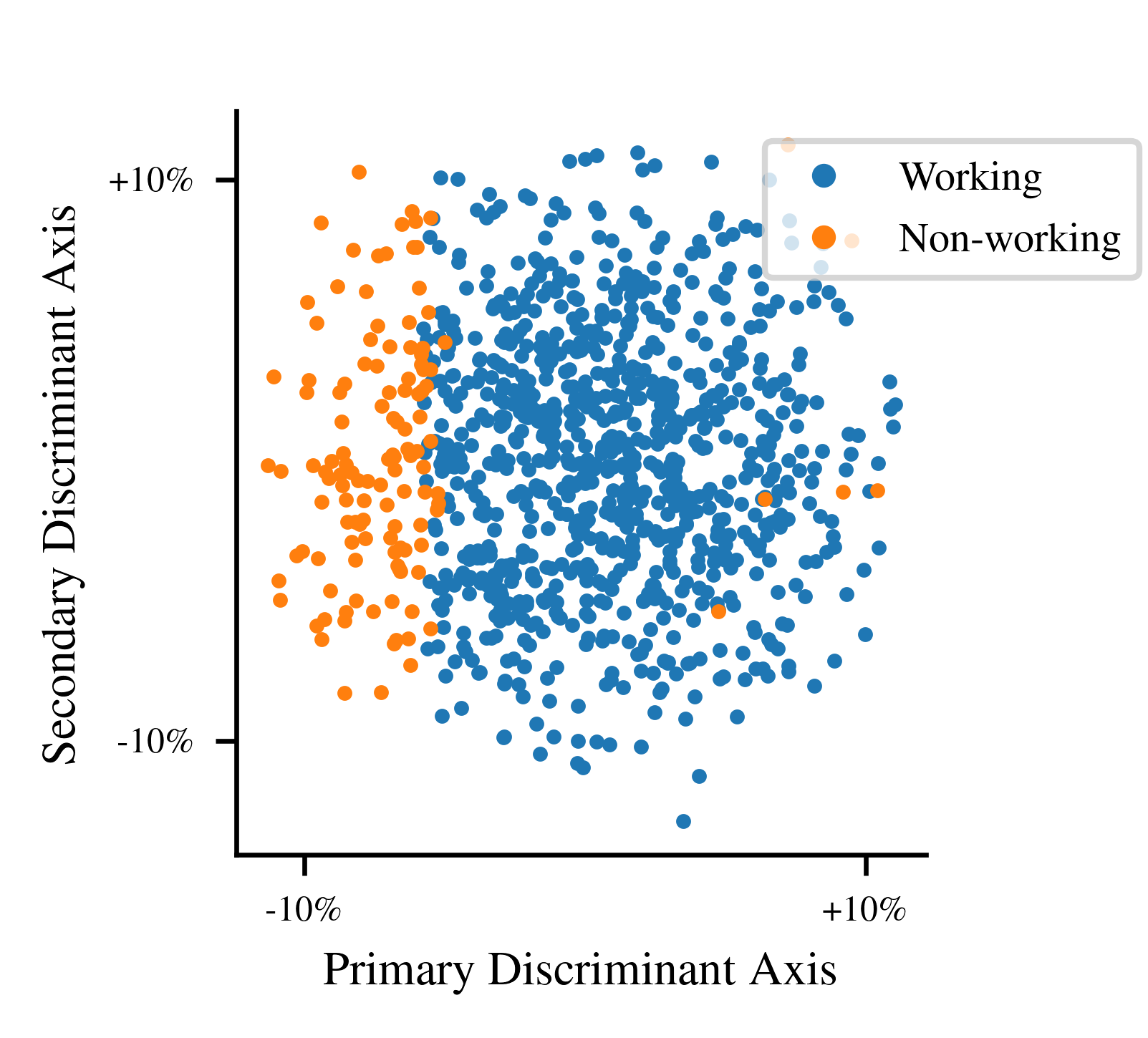}
    \caption{\textbf{Parameter variation point cloud.} A point cloud
        of the two-dimensional projection \edited{by LDA (described
        in the text)} of the twelve-dimensional variable $\xi$
        representing fractional variation in parameter values
        (increased to 15\% total for illustrative purposes) for 1000
        realizations of the CPG module. Blue points represent
        parameter values under which the module continues to function,
        and orange points represent those for which a failure was
        detected. The orange points in the majority-blue region are a
        result of the projection to lower dimension, rather than of
        inconsistent or nondeterministic behavior.
    } \label{f:pcacloud}
\end{figure}

Indeed, \edited{the point cloud is nearly linearly separable, i.e.}\
nearly every module failure can be attributed to the component of the
deviation which occurred along a single axis. In this direction, a 5\%
deviation was sufficient to cause a failure, whereas even 15\%
deviation along other axes did not have this effect. The projection of
the higher-dimensional point cloud onto a two-dimensional plane is
shown in \cref{f:pcacloud}. \edited{The horizontal axis of the
projection is the discriminant axis returned by LDA, but the vertical
axis is arbitrary because all remaining directions are equally
(un)informative.}

This linear discriminant axis is
aligned with changes in the threshold, resting, and reversal potential
parameters $V_t$, $V_r$, and $V_n$. Sensitivity to these  parameters is
not necessarily significant to the system design, however, because
voltages in biological systems are tightly controlled through homeostatic
regulation of ionic concentrations \cite{hille1992ion}. Parameters
such as conductance which are not directly regulated are typically
expected to vary much more with temperature \cite{rinberg2013effects};
these parameters have much less effect on our system behavior.

Although the points in this projection are almost linearly separable
along this axis, a few points within the larger cloud correspond to a
module failure but appear to be outliers because the projection has
moved them away from the border of the allowed region of parameter
space. In our system, just as has been observed in computational
models of a lobster digestive CPG \cite{marder2006variability}, the
allowed region of parameter space has nontrivial higher-dimensional
geometry. \edited{In particular, the same behaviors could have been
achieved using entirely different parameter values corresponding to
other cell types.} We observe near-perfect linear separability because
our initial parameter  values are relatively close to a particular
failure mode, but as the size of the point cloud increases, these
lower-dimensional projections become  progressively less informative
as they become dominated by apparent outlier  points, which make up
only a small fraction of the points in \cref{f:pcacloud}.

\section{Robotic Case Study} \label{s:casestudy}

The basic motion primitives underlying locomotion are the periodic
oscillations which drive different motor components. For example, the
\celegans{} locomotor system is based on a central pattern generator
which produces a repeating ripple pattern down the body of the worm,
propelling it forward~\cite{zhen2015elegans}. Likewise, in many
insects, each leg is associated with a CPG that produces a single step
and can be modulated by higher-level control
\cite{bidaye2017sixlegged} as well as sensory feedback
\cite{berendes2016speeddependent} to generate the full walking gait of
the organism.

We can design a central pattern generation network of this type for a
robotic application by combining the abstraction of asynchronous
digital logic with analog effects inspired by these biological
prototypes. The remainder of this paper will be focused
on the design and validation of a central pattern generation network
for a flexible modular robot. This robot produces a forward crawling
gait, \edited{which can be viewed as a model of worm locomotion,}
using four linear actuators driven by a simple four-phase finite state
machine.

\subsection{The Robot and its Gait} \label{s:robot}

\begin{figure}
    \centering
    \includegraphics{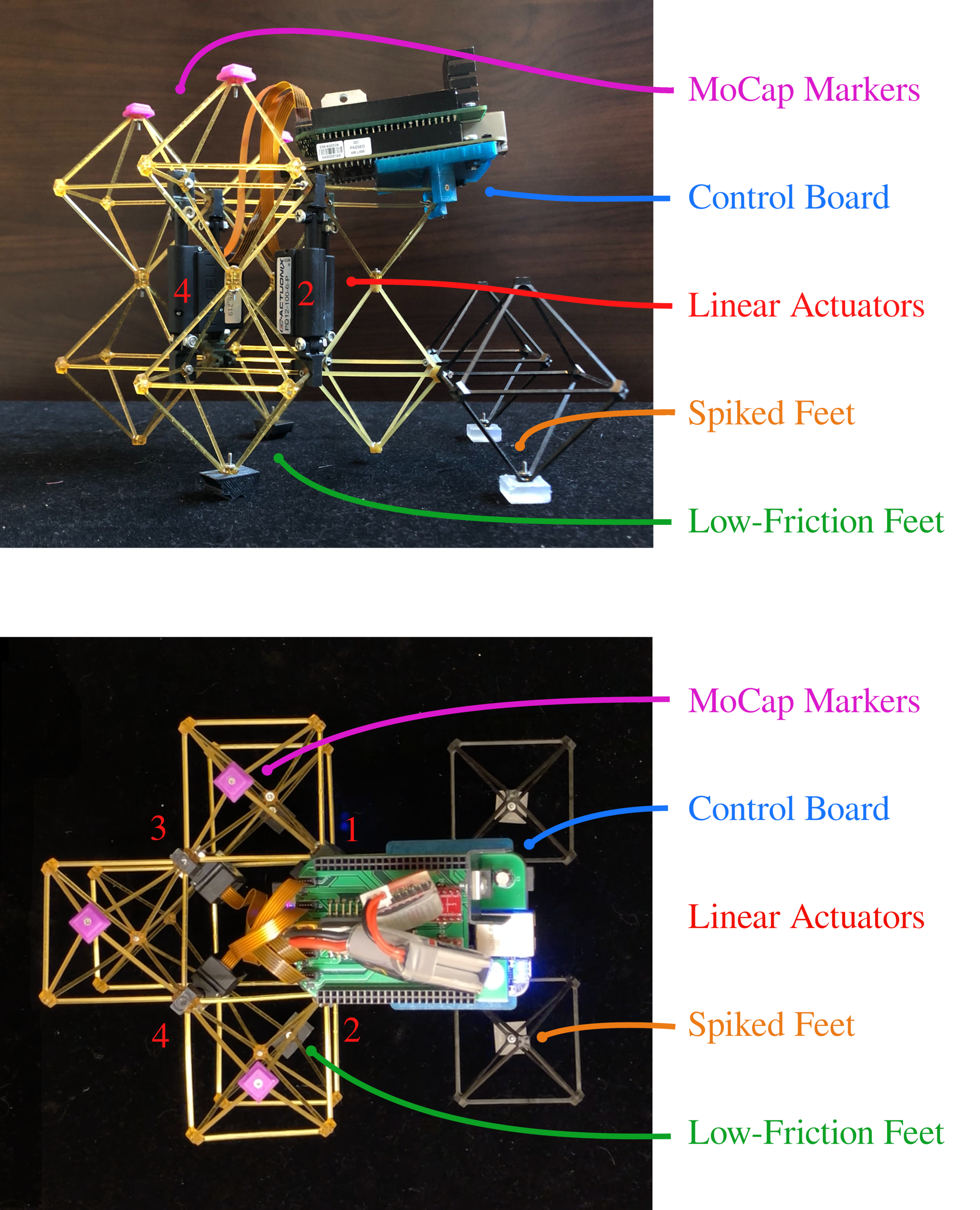}\hspace{-100pt}
    \caption{\textbf{The physical robot.} A diagram of the flexible
        modular robot for which our spiking central pattern generator
        is designed, shown in both side-on (above) and top-down
        (below) views. The three different voxel colors correspond to
        the three different materials described in the text.
    } \label{f:voxelbot}
\end{figure}

The ideal robotic application for a minimal CPG is one whose
locomotion is produced by rhythmic activity, but which does not need
high-bandwidth control logic to stay upright. Past researchers have
fulfilled this requirement with a variety of different statically
stable robotic systems, in particular many different types of modular
robots \cite{vonasek2015highlevel, fan2016chaotic, wang2018rhythmic,
chen2019braininspired}. Likewise, the robotic application for our
CPG is a modular robot constructed from volumetric pixels
\cite{cramer2017design}. These ``voxels'' take the form of cubic
octahedra, which can be interlocked using nuts and bolts to create a
deformable lattice that in larger structures can function as a
metamaterial \cite{cheung2013reversibly}. Ten voxels are arranged into
two layers in the configuration shown in \cref{f:voxelbot}, with
3D-printed feet attached to four of the voxels on the bottom layer.

The robot electronics comprise four DC linear actuators and a control
module containing custom motor drive electronics as well as a
BeagleBone Black single-board computer. We use DC linear actuators
rather than the linear servos which are common in similar applications
because servos do not provide pose information to the software. Each of
the four actuators is controlled through an L298 DC motor driver,
which receives a motor direction signal and PWM enable input from the
microcontroller. Each actuator also contains a linear potentiometer
used as a voltage divider to produce a voltage proportional to the
actuator position; the ADC reads a 10-bit fixed-point number equal to
the actuator extension as a fraction of its stroke. The CPG network
can use this information as a form of proprioceptive feedback in order
to implement closed-loop control rather than relying on the position
control built into a servo.

Our robot's walking gait is generated by four distinct, symmetrical
actuation phases, each of which requires the robot to extend and
contract two actuators which mirror each other while the other two are
held in place. First, actuator \#1 extends while actuator \#3
contracts. This lifts the left front foot off the ground, while the
right front foot remains planted. Second, actuator \#2 extends while
actuator \#4 contracts; this deforms the robot's body so that the left
front foot, which is no longer in contact with the ground, moves
forward. During this motion, the left front foot remains in place on
the ground due to the friction between the surface and the spiky
3D-printed foot, while the smooth rear feet are free to slide. Next,
actuator \#1 contracts while actuator \#3 extends, lowering the left
front foot and raising the right. Finally, the right foot is moved
forwards by contracting actuator \#2 while extending actuator \#4.

An advantage of the modular physical design through voxels is that
individual structural subunits can be swapped out for others with
different material properties---for example, our robot uses three
different voxel materials. The main body is constructed from the
most compliant resin so that it does not present significant
resistance to the actuators, while the front feet are made from
relatively rigid carbon fiber for more efficient use of the
deformation generated in the body. Additionally, the node on the
bottom to which the feet attach is made from a slightly stiffer
resin in order to transmit motion to the feet more effectively.

\subsection{Development of the CPG}

We directly implement a state machine which controls the actuation of
the original robot; four neural \edited{latch} modules connected in a
ring represent a one-hot encoding of the current state. Once any of the
four modules is activated, for example by an external input, it
gradually activates its successor, which in turn deactivates the
first module. \edited{The result is a new form of
oscillation; rather than the simple limit cycle of alternating firings
produced by the original module, each module undergoes
regular alternation between active and inactive states. Although the
dynamical mechanism is not strictly bursting, similar bursts of spikes
surrounded by quiescence are produced.}

\begin{figure}
    \centering
    \includegraphics{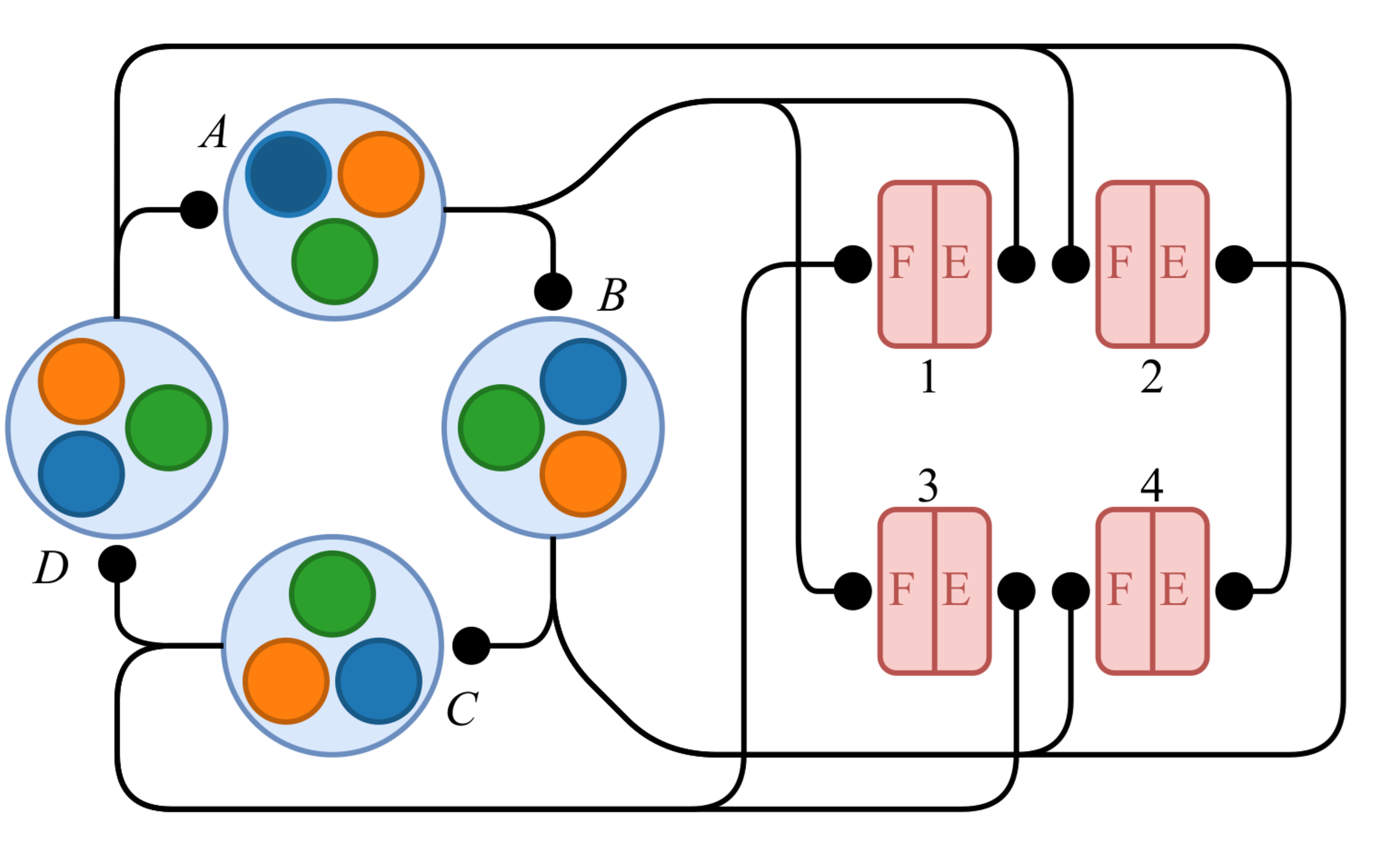}
    \caption{\textbf{The full CPG network.} The system consists of
        four individual modules A--D connected in a ring topology.
        Each module is associated with one flexor (F) and one extensor
        (E) among the 8 muscle cells which directly control the DC
        actuators. The pairs of muscle cells are labeled 1--4 in
        correspondence with the physical actuators. Note that the
        depicted connections between modules are actually
        bidirectional---a given module activates its successor while
        activating the inhibitory interneuron of its predecessor.
    } \label{f:network}
\end{figure}

If we label the four modules A--D and identify them
with the four states of our state machine, the system simply transitions
through the states in alphabetical order.
This corresponds to the four actuation phases
described previously, where at any given time during the operation of
the robot, one of the four actuators is expanding while another is
contracting. At the same time, the other two actuators remain fixed in
place in the absence of new motor input due to the mechanical
properties of the worm gear rack-and-pinion actuation. As a result,
with a state machine that cycles through four states, the desired
pattern of muscle actuation is generated simply by having each state
excite the corresponding extensor and the opposite flexor. The result
is the full network depicted in \cref{f:network}.

This network implements the correct state machine, but the
speed with which it switches between states is parameter-dependent and
much too fast to produce the desired gait. The inertia and damping
inherent to the mechanical construction of the robot mean that the
actuators simply oscillate in place without producing any useful
forward motion. For this reason, it becomes necessary to use feedback
to modulate the speed with which the network switches between states.
One can imagine that, as in \celegans{} \cite{zhen2015elegans}, the
neurons of our CPG are equipped with strain-sensitive neurites which
create an inhibitory leakage current in proportion to the deviation of
the preceding module's actuator from its target position. This
prevents the initiation of the next state when the current state has
not yet achieved its intended actuator position.

In order to describe the proprioceptive feedback $I_{\text{prop},i}$
to the $i$th module due to this linear feedback current, we introduce
the actuator position variables $z_i$ and $z_{i*}$ representing the
relative extension of the previous actuator and its opposite
counterpart as numbers between 0 and 1. Additionally, there is a
feedback constant $k_p$ which must be tuned by a bit of trial and error
but which depends only on the neuron parameters, not the physical
properties of the robot. In our application, a value $k_p =
\SI{25}{\pA}$ was effective. The resulting feedback current is given
by:

\begin{equation}
    \label{eq:feedback}
    \begin{aligned}
        I_{\text{prop},i} &= -k_p\abs{1 - z_i} - k_p\abs{z_{i*}}
                       \\ &= -k_p(1 + z_{i*} - z_i)
    \end{aligned}
\end{equation}

\begin{figure}
    \centering
    \includegraphics{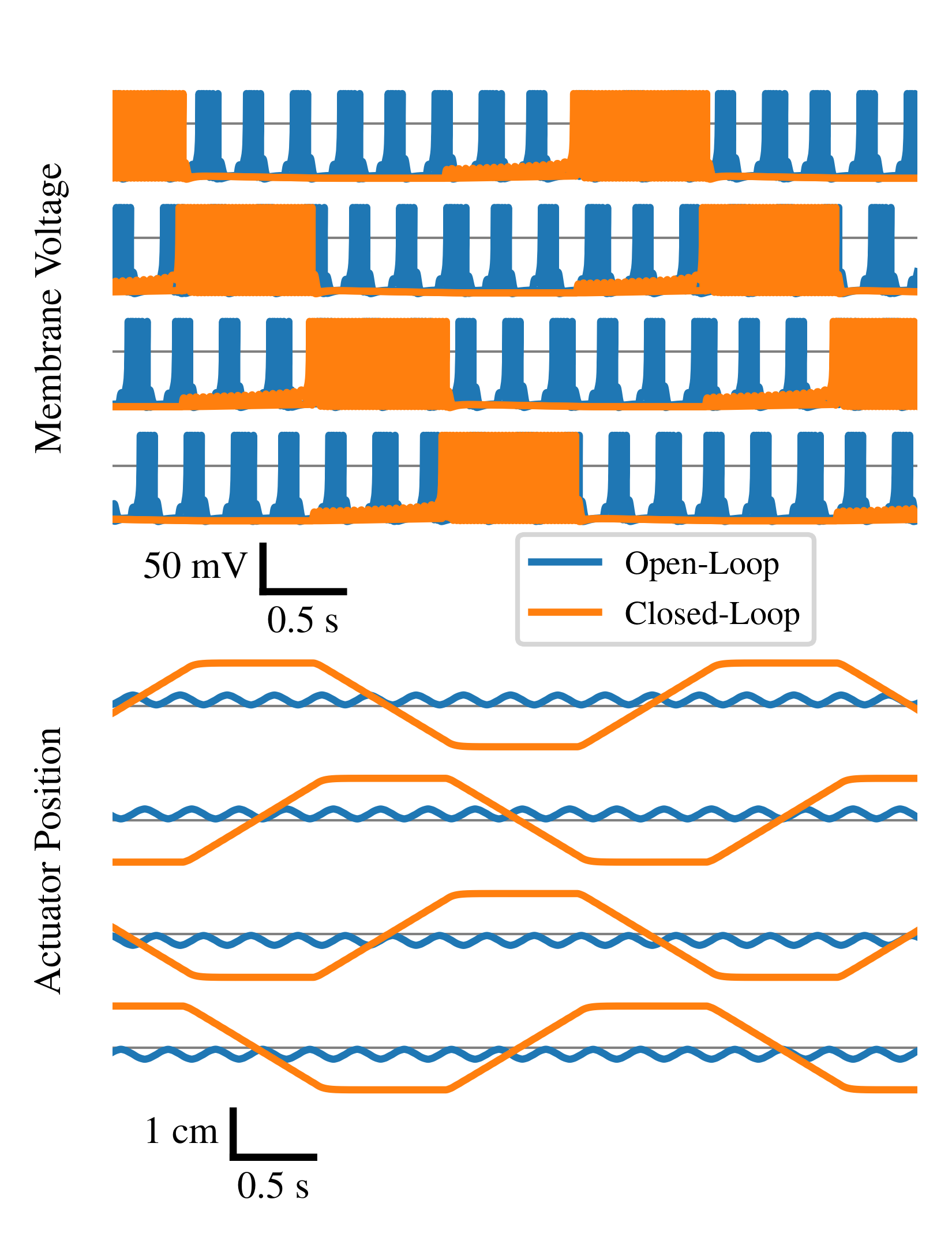}
    \caption{\textbf{Open vs.\ closed loop.} Comparison in simulation
        between the open-loop and closed-loop oscillatory behavior of
        the designed neural network. The top four traces represent
        overall neural activity in each of the four modules during
        the two experimental conditions, and the lower four traces
        represent the extension of the actuators.
    } \label{f:comparison}
\end{figure}

The behavior of the open-loop and closed-loop networks are contrasted
in \cref{f:comparison}. Both network configurations produce the same
pattern of alternation between states as well as the same pattern of
actuation, but the short duration for which each state remains active
in the open-loop network means that the actuators only extend a very
short distance before being forced to contract again. Note that this
demonstration was performed in simulation for clarity; the depicted
actuator position is derived from treating the actuator kinetics as
ideal linear damped masses. In the realistic case, the difference is
more extreme because frictional losses cause the open-loop excursion
of the actuators to be significantly less than pictured here. As a
result, the robot under open-loop control moves extremely
inefficiently or not at all.

\section{Physical Experiment} \label{s:physical}

To this point, all description of the full spiking neural network
depicted in \cref{f:network} has been simulated, qualitative, or both.
In order to provide a concrete physical demonstration of the efficacy
of our approach, in this section we describe the experiments which
were carried out in order to validate the physical robotic platform
just described.

\begin{figure}
    % This image is too wide for the textwidth here, but within PLoS
    % requirements, so manipulate the margins with ad-hoc spacing.
    % \makebox[\textwidth]{\hspace{-45pt}\includegraphics{Fig8}}
    \hfuzz 130pt\includegraphics{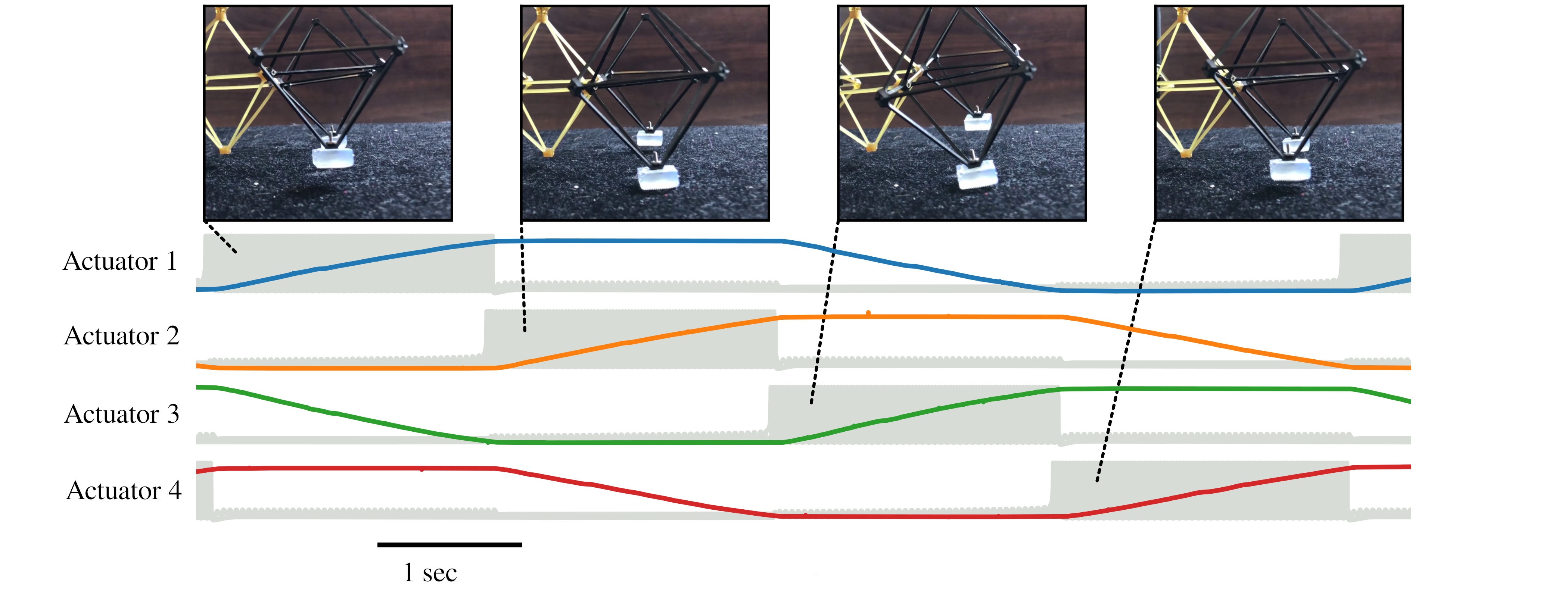}
    \caption{\textbf{Robot gait time-lapse.} A time-lapse of the
        motion of the robot's front feet, overlaid on top of an
        abstract representation of the neural activity  necessary to
        produce the depicted motion. The gray traces are overlays of the
        membrane voltage of all three neurons in each module, which
        disappear into each other due to the long timescale, and the four
        colored traces
        represent the extension of the four actuators. The four
        photographic insets correspond to the four phases of actuation
        described in the text. First, the right foot is raised above
        the ground as it moves forward. Next, the right foot comes
        down and both feet are briefly planted. Then, the left foot is
        raised and begins to move forward. Finally, the left foot is
        lowered to the ground again, and the cycle continues.
    } \label{f:timelapse}
\end{figure}

During all of these physical experiments, we recorded the full state
of the neural network as well as the commanded actuator effort and
outputs of the proprioceptive sensors. This enables us to directly
correlate the activity of the neural network with the physical
behavior of the robot, as in \cref{f:timelapse}.

In order to quantify the physical movement of our robot, we
implemented a simple motion capture system through color-based
particle tracking, with three magenta 3D-printed markers affixed to
the top of the robot. We assume that once the camera is calibrated to
minimize distortion, position in the image plane is proportional to
physical position in the ground plane. This assumption allows us to
roughly calibrate the system using the fact that the distance between
the motion capture markers while the robot is in a neutral pose is
determined by the known geometry of the voxels.

\begin{figure}
    \centering
    \includegraphics{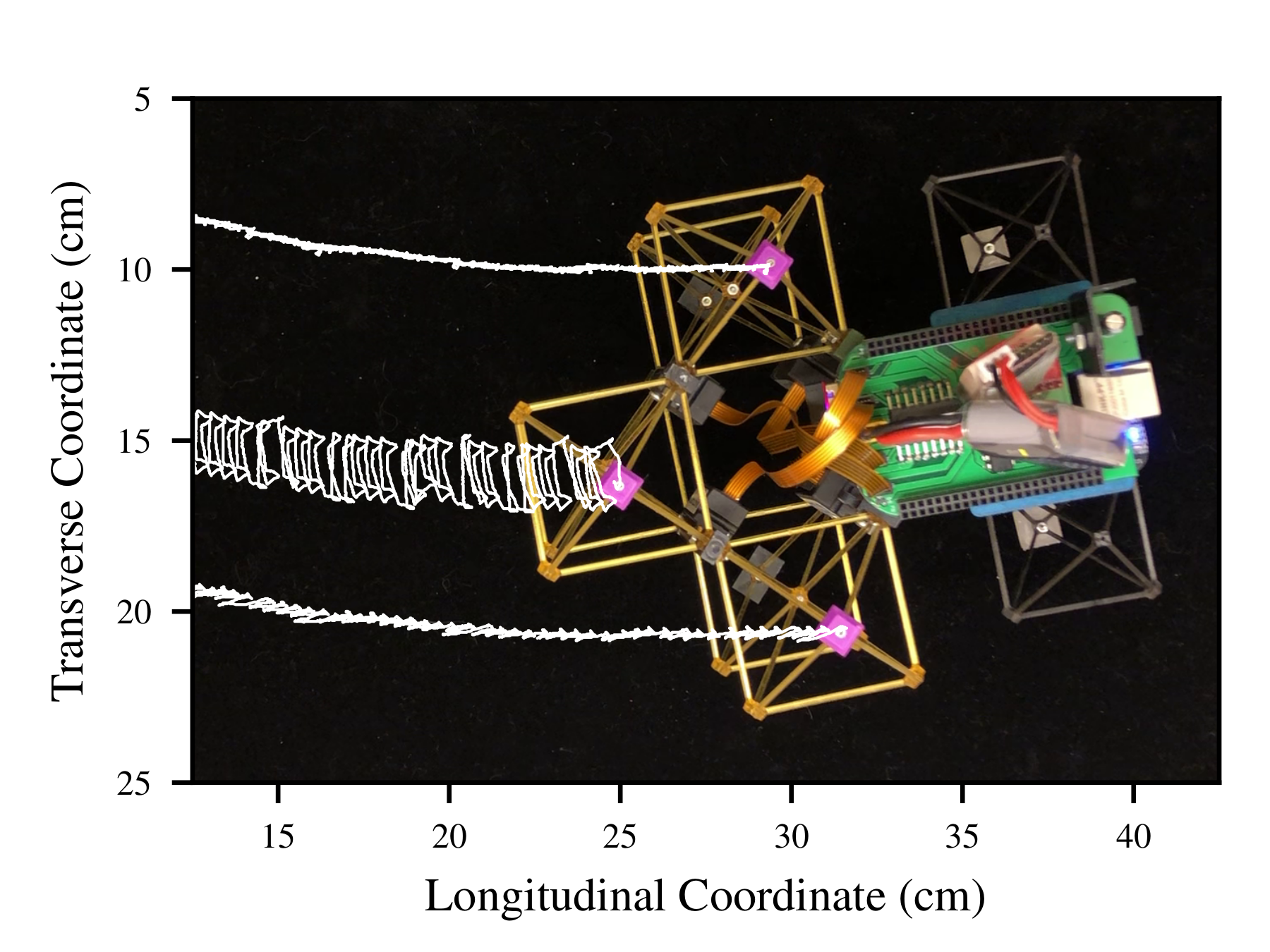}
    \caption{\textbf{Overhead motion capture.} Motion capture traces
        demonstrating the movement in the horizontal plane (from right
        to left) of three plastic motion capture markers affixed to
        the top of the robot. White traces represent the position of
        the purple markers over approximately five minutes.  The
        oscillatory patterns in the white traces are not noise in the
        motion capture system, but rather represent the cyclic
        trajectory traced out by the voxel nodes.
    } \label{f:mocap}
\end{figure}

\begin{figure}
    \centering
    \includegraphics{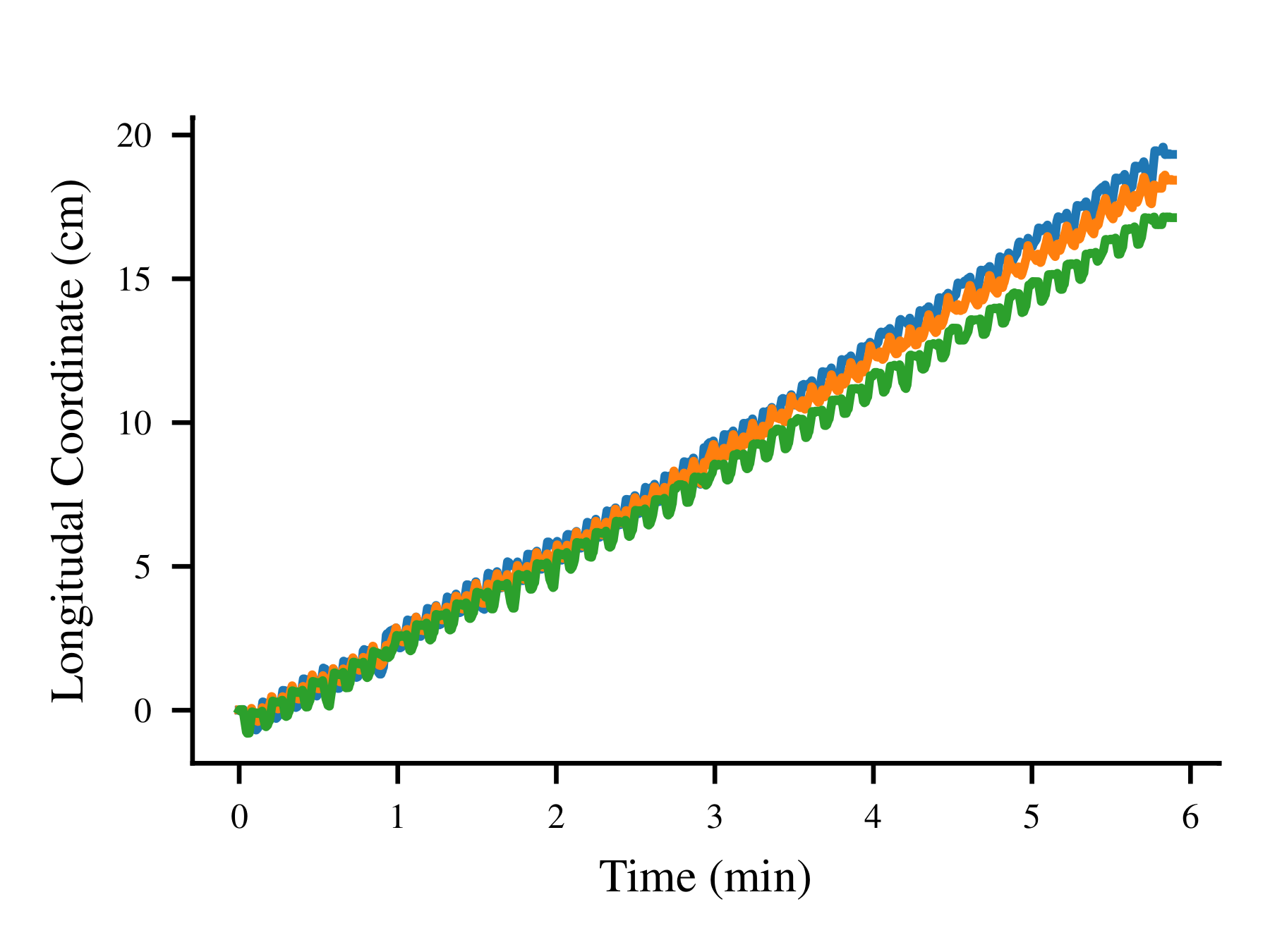}
    \caption{\textbf{Forward motion of the robot.} The forward
        displacement of the three motion capture markers from their
        starting positions during the motion displayed schematically
        in \cref{f:mocap}.
    } \label{f:forward-x}
\end{figure}

An example of the type of data which can be captured by this system is
presented in \cref{f:mocap}, where the motion of the markers during
forward locomotion is plotted on top of an image of the robot's
position at the end of the recording. The physical position of the
robot during this behavior is shown in \cref{f:forward-x}; with
this motion-tracking data, we can approximate the robot's forward
speed to be \SI{3}{\cm\per\min}, mainly limited by the slew rate of
the actuators.

Next, we implemented a variation on our controller which is capable of
reversing direction in response to an aversive stimulus. In a simple
network like this, the easiest way to implement two different
locomotive directions is the route taken by \celegans: introducing what
amounts to two parallel copies of the same CPG network
\cite{zhen2015elegans}, one of which is active during forward movement
and one of which is active when the robot is moving backward. We then
introduce a single neuron responsible for detecting the presence of
some aversive stimulus and signaling that the forward network should
be deactivated and the reverse network activated.

\begin{figure}
    \centering
    \includegraphics{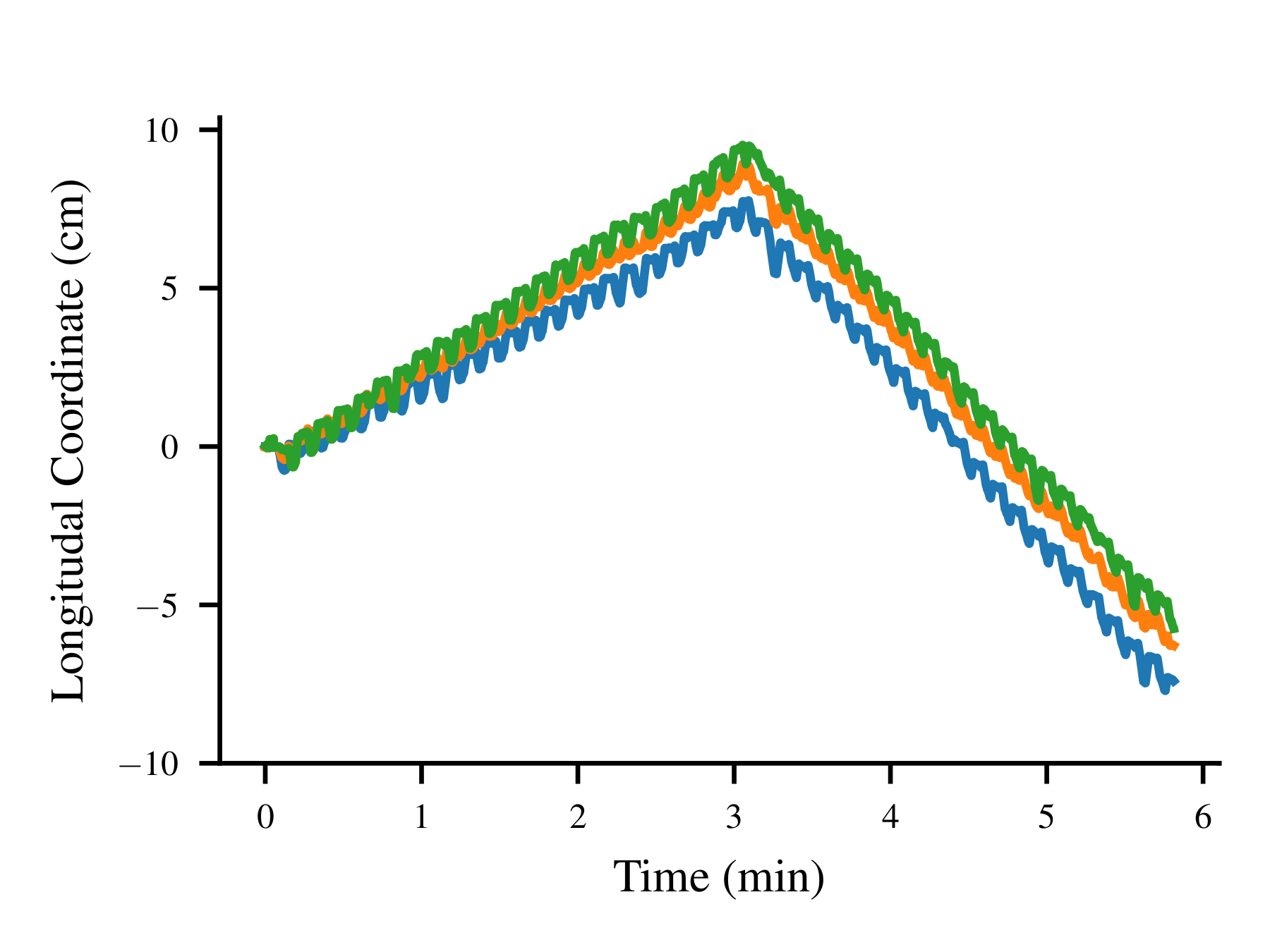}
    \caption{\textbf{Reflexive reversal of the robot.} The forward
        displacement of the three motion capture markers from their
        starting positions during the reversal reflex experiment.
    } \label{f:backward-x}
\end{figure}

The result is that as soon as an aversive stimulus is detected during
forward motion, the robot reverses direction and begins to move
backward instead. In our experiments, the ``aversive stimulus'' of
choice occurred at an arbitrary time, but it would be simple to
replace this behavior with something more reactive, such as a
physical bumper to detect obstacles. The position of the robot over
time during this experiment is shown in \cref{f:backward-x}.

\section{Conclusion}

We have described a simple approach to the design of spiking
neural networks for robotic control. By taking advantage of the fact
that the behavior of neurons can be approximated by a simple binary model
we can design a connectome capable of producing the desired behavior.
A case study was provided by a flexible modular robot, where we applied
this approach to produce a tripod-stable gait which can be modulated by
sensory feedback or external inputs much like a biological central pattern
generator.

In the future, we are interested in developing a method to efficiently
characterize the geometry of the extended region of parameter space in
which the designed neural module continues to function. This would
allow a more thorough approach to the question of robustness vs.\
modulation, as the acceptable region of parameter space is clearly
significantly larger and more irregularly shaped than the region
studied here. \edited{It would also be of interest to more rigorously
study the question of how single-module robustness translates to the
full system.}

Additionally, in the real-world evolution of central pattern
generation networks, although the simple underlying architecture tends
to be evolutionarily conserved, evolution tends to provide a great
deal of variety in the ways that this architecture can be modulated by
higher-level control \cite{katz2016evolution}. We seek to explore
such additional forms of modulation in future work; for example, we may
introduce a higher-level learned or evolved network which interacts
with the world by modulating or controlling an underlying conserved
central pattern generation network.

\section*{Supporting Information}

\paragraph{S1: Simulation Code}
A Jupyter notebook containing the Julia code which implements our
simulations as well as the mathematical results of
\cref{s:robustness}.

\paragraph{S2: Forward Locomotion}
A video of the robot performing forward locomotion, from which the
motion capture data in \cref{f:mocap,f:forward-x} were computed.

\paragraph{S3: Backward Locomotion}
A video of the robot performing backward locomotion, from which the
position data depicted in \cref{f:backward-x} were computed.

\paragraph{S4: Side View}
Side view of the robot taking a few steps, used to generate
\cref{f:timelapse}.

\section*{Acknowledgments}

The authors would like to acknowledge the technical support of the
Braingeneers research group as well as a donation made possible by
Eric and Wendy Schmidt by recommendation of the Schmidt Futures
program.

\end{document}